\documentclass[10pt,twocolumn,letterpaper]{article}

\usepackage{cvm}
\usepackage{times}
\usepackage{epsfig}
\usepackage{graphicx}
\usepackage{amsmath}
\usepackage{amssymb}
\usepackage{booktabs}

\newcommand{\jdq}[1]{{\color{black}#1}}
\newcommand{\zzhu}[1]{{\color{black}#1}}
\newcommand{\zy}[1]{{\color{black}#1}}

\usepackage[pagebackref=true,breaklinks=true,letterpaper=true,colorlinks,bookmarks=false]{hyperref}

\cvmfinalcopy 

\ifcvmfinal\fi
\begin{document}

\title{Sphere Face Model:A 3D Morphable Model with Hypersphere Manifold Latent Space}

\author{Diqiong Jiang\\
Zhejiang University\\
\and
Yiwei Jin \\
Zhejiang University\\
\and
Fang-Lue Zhang\\
Victoria University \\
of Wellington\\

\and 
ZHE ZHU \\
Duke University\\

\and
Zhang Yun \\
Communication University \\
of Zhejiang \\
\and
Tong, Ruofeng\\
Zhejiang University\\
 
\and
Tang, Min \\
Zhejiang University\\
}

\maketitle
\begin{figure*}[t]
\includegraphics[width=\linewidth,trim={0.5cm 4.5cm 3cm 5cm},clip]{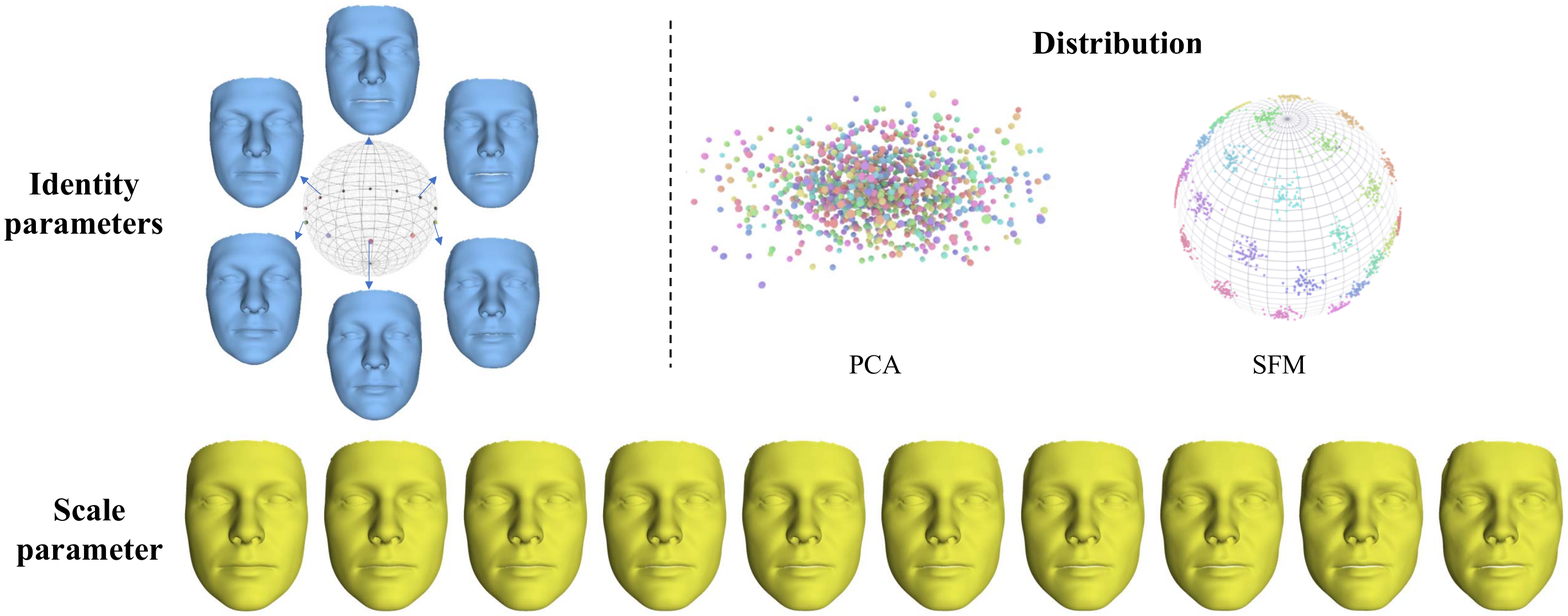}
\centering
\caption{The overview of the sphere face model. (a) The identity parameter of the sphere face model distributed on the hypersphere represents the identity information.  The meshes are uniformly sampled from on the hypersphere using the first two dimensions of identity parameters. (b)The scale parameter of the sphere face model is scalar, which controls the distinctiveness to the mean face. (c)The distribution of the parameter. The shape parameter of the PCA-based model has an anisotropic multivariate Gaussian distribution. Our identity parameters are distributed isotropically on the hypersphere and separated between different classes.}
\label{fig:teaser}
\end{figure*}

\newcommand{\jyw}[1]{{{\color{blue}{#1}}}}
\begin{abstract}
3D Morphable Models (3DMMs) are generative models for face shape and appearance. However, the shape parameters of traditional 3DMMs satisfy the multivariate Gaussian distribution while the identity embeddings satisfy the hypersphere distribution, and this conflict makes it challenging for face reconstruction models to preserve the faithfulness and the shape consistency simultaneously. 
To address this issue, we propose the Sphere Face Model(SFM), a novel 3DMM for monocular face reconstruction, which can preserve both shape fidelity and identity consistency. The core of our SFM is the basis matrix which can be used to reconstruct 3D face shapes, and the basic matrix is learned by adopting a two-stage training approach where 3D and 2D training data are used in the first and second stages, respectively. To resolve the distribution mismatch, we design a novel loss to make the shape parameters have a hyperspherical latent space.
Extensive experiments show that SFM has high representation ability and shape parameter space's clustering performance. Moreover, it produces fidelity face shapes, and the shapes are consistent in challenging conditions in monocular face reconstruction.

\end{abstract}
\section{Introduction}
The problem of face reconstruction from stills and videos has been attracting considerable attention in the computer vision and computer graphics community. It has a broad range of applications, including AR/VR\cite{chen2018real}, animation\cite{lattas2020avatarme}, computer games\cite{lin2021meingame}, etc. In recent years, there is a growing demand for customizing 3D virtual faces for game characters\cite{lin2021meingame,shi2020fast} or personalized 3D facial editing\cite{yang2021learning}. In such applications, images from common users usually come from a large diversity of conditions, including occlusion, resolution, pose, expression, illumination, etc. It is thus challenging to reconstruct a face from only a single image requiring both shape faithfulness and identity preservation. 

\jdq{Although previous works\cite{zhu2017face,jourabloo2016large} claimed to have achieved face reconstruction from a single image, their reconstructed face shapes suffer from inconsistent identity properties when the input images have varying conditions.}
To solve this problem, the follow-up works\cite{sanyal2019learning,tuan2017regressing,liu2018disentangling} propose to aggregate shape parameters
of the same identity \zzhu{while} separate those of different subjects to produce 3D face shapes containing good identity-related features. 
However, the conflict between the shape loss and the identity loss in their reconstruction pipeline prevents them from achieving both shape fidelity and identity consistency.
That conflict comes from the \zzhu{mismatch} between the distribution of identity embeddings of face recognition and shape parameters of the previous 3DMMs \cite{paysan20093d,li2017learning,gerig2018morphable,booth20163d}, which maximize their model expression ability while neglecting some distinguishable information of categories.  


Therefore, this paper focuses on \zzhu{identity-consistent face reconstruction in a linear model. We resolve the distribution mismatch problem and propose }a novel \zzhu{face generation} model called the Sphere Face Model (SFM). We add category information while building the basis of SFM, and constrain identity parameters over a hypersphere by normalizing the shape parameters to make the latent space of SFM consistent with identity embeddings. In this way, we resolve the conflict between the two losses and further improve the identifiability of 3D face models. Moreover, SFM has an essential property that the discrimination of the parameters is transferable to the geometry, which means \jdq{the Euclidean distance between of two sets of 3DMM parameters in parameter space and between corresponding mesh vertices in geometry space have a positive correlation.}  \zzhu{One notable challenge is }  when the identity parameters are forced to be distributed over a hyperspherical surface, the \zzhu{L2 norm of the parameter vectors become the same. In another word, the reconstructed faces would have the same root mean square errors from the mean face, leading to reduced varieties of generated faces. We use two approaches to address that issue. Algorithmically we add a parameter to control the scale of the shape parameters of each face. While previous approaches mainly use 3D training data which are limited, we propose a two-stage training approach where we use 3D data only for pre-training and  adopt  an unsupervised learning approach that can leverage a sufficient amount of 2D face data.} 
Figure \ref{fig:teaser} \zzhu{highlights the differences }between our face model and the previous 3DMMs. The parameter of SFM is composed of an shape parameter and a scale parameter. The identity parameter is the normalized shape parameter, which controls the face's identity attribute. It is distributed on the hypersphere with good separation properties. The scale parameter controls the distance to the average face. 
 
The main contributions of this paper lie in the following three aspects: 
\begin{itemize}
\item We propose Sphere Face Model (SFM) for 3D face reconstruction from single images with \zzhu{both} shape faithfulness and identity consistence. 
\item We propose a new structure of 3DMMs, where the latent space follows a hypersphere distribution and the discrimination of parameter space is transferable to the geometry space. 
\item To enable SFMs to reconstruct high-quality 3D face models from single images, we present a learning scheme to train SFMs with both 2D and 3D data.
\end{itemize}


\section{Related work}
3D morphable models map the high-dimensional face geometry space to the low-dimensional manifold space. \jdq{Based on 3DMMs, the previous works optimize the low-dimensional 3DMM parameters from the input image to reconstruct high-dimensional face geometries in monocular face reconstruction. Meanwhile, many works introduce identity loss in the face reconstruction pipeline to keep the face shape stable from the various input image in recent years. } This section introduces the related works from three aspects: \jdq{3D morphable model, shape-consistent face reconstruction from monocular images, and deep face recognition. }



\textbf{3D morphable models}
3D morphable model is a statistical model of the distribution of the faces, which maps the low dimensional parameter vector to the high dimensional graphic vertices. The groundbreaking work of 3DMMs traces back to Blanz and Vetter\cite{blanz1999morphable}, who propose the 3D morphable model using principal component analysis from an example of 200 3D faces. Based on this idea, Paysan et al.\cite{paysan20093d} provide the first public 3DMM model, BFM 2009 and others  \cite{blanz2003reanimating,thies2015real,amberg2008expression,li2010example,bouaziz2013online,vlasic2006face} extend the model to introduction emotive facial shapes information by adopting an additional expression basis or using bilinear and multilinear. \cite{li2017learning} provides the whole head model, FLAME, which introduces an articulated jaw, neck, and eyeballs in linear shape space and global expression blendshapes to make the model more expressive. Unlike the previous work, we consider identity information while constructing the 3DMM model, and the shape parameter can be inherently separated among each identity. Blanz and Vetter\cite{blanz1999morphable} only use facial meshes of 200 subjects of similar ethnicity and age, which cannot represent the great diversity of the human faces. \cite{booth20163d} train the \zy{3DMM} with the large scale of 3d data to overcome this limitation, but the 3D data is also limited. \cite{tran2018nonlinear,tewari2018self,tran2019towards} use sufficient 2D data to training the 3DMM. \jdq{However, training with 2D data without 3D prior need strong regular term, which leads to lack of geometric details and diversity.} Our method training the model make full of 2d and 3d data. In recent years, with the development of deep  learning,\cite{tran2018nonlinear,tran2019towards,bagautdinov2018modeling} propose nonlinear models with encoder-decoder structure. \jdq{Those nonlinear models do not consider the parameter separation and the property of propagate the discrimination from parameter space to geometric space when training the models.}

\textbf{Shape consistence monocular face reconstruction } 
Early works\cite{aldrian2011inverse,schneider2017efficient,bas2016fitting,paysan2009face,schonborn2017markov}, reconstruct 3d face from monocular RGB using the analysis-by-synthesis approach with the prior knowledge of the 3DMM. They often apply the photometric and landmark consistency between the input and the rendered image. In recent years, many \zy{researches}\cite{guo2020towards,sanyal2019learning,deng2019accurate} have proposed the deep network to regress the 3DMM parameters. 
Adversarial loss, perceptual loss, and identity loss on the rendered image\cite{lin2020towards,gecer2019ganfit,tran2019towards} \zy{are proposed} to generate the high fidelity texture. To \zy{reconstruct} the stable face shape geometry, Tran et al. \cite{tuan2017regressing} label a large number of face images with 3DMM shape parameters using \zy{the} optimization method, and \zy{utilize the} deep CNN to learn the mapping from images to shape parameters. 
Liu et al.\cite{liu2018disentangling}, and Sanyal et al.\cite{sanyal2019learning} use a face recognition loss to push away the shape parameters of different people while aggregating those of the same person. Jiang et al.\cite{jiang2021reconstructing} explore the relationship of shape parameter discrimination and geometric visual discrimination and proposal the SIR loss, which increase discriminability in both the shape parameter and shape geometry domain. \jdq{Their methods do not focus on the properties of the 3DMM shape basis, which is also a crucial factor of shape consistence in monocular face reconstruction.}

\textbf{Deep face recognition}
Many works have achieved incredible face recognition accuracy with the powerful deep convolutional neural network in recent years. Most of them focus on cleaning and mining the training data, or designing the loss function to maximize the intraclass distance and minimize the interclass distance, which boosts the discrimination of deep feature identity embedding. There are mainly three types of loss function for face recognition. One utilizes pair or triple training strategy, such as contrastive loss \cite{sun2014deep} and the Triplet loss \cite{schroff2015facenet}. Another type of loss, like \zy{the} center loss\cite{wen2016discriminative}, plays as the auxiliary loss to augment the other loss functions. The aim of these loss functions is aggregating features to \zy{minimize} the inner-class distance. The auxiliary loss can be directly added to the classifier network and learn the discriminative features. The last type of loss is modified softmax\cite{liu2016large,wang2017normface,liu2017rethinking,liu2017sphereface,deng2019arcface,wang2018cosface}. Normface\cite{wang2017normface} and Cocoloss\cite{liu2017rethinking} normalized the weights and features and directly optimize the cosine similarity instead of the inner product. L-softmax\cite{liu2016large} and sphereface\cite{liu2017sphereface} introduce the multiplicative cosine margin. Cosface\cite{wang2018cosface} and Am-softmax\cite{wang2018additive} introduce the additive cosine margin, and arcface\cite{deng2019arcface} introduce\zy{s} the additive angular margin. 
\cite{huang2020curricularface,zhang2019adacos,liu2019adaptiveface} adapt the margin during the training. Current SOTA deep face recognition methods mostly adopt the last type of loss, and softmax-based classification loss. Their identity embedding space is the hypersphere.
\section{Latent Space Distribution}
In this section, we elaborate the characteristics the latent space need to have for effective face representation and reconstruction. 

Previous 3DMM-based works have a conflict between the losses for face recognition and reconstruction in the shape-consistent face reconstruction pipeline. Taking PCA-based face models as an example, the shape parameters for face reconstruction satisfy the anisotropic multivariate Gaussian distribution. In contrast, the identity embeddings for face recognition are distributed isotropically on the hypersphere. \jdq{The distribution mismatch in the latent space of face reconstruction and face recognition makes the co-convergence of these two loss functions (face recognition loss and face reconstruction loss) very difficult to achieve.}
More specifically, when conducting the intense face recognition loss, the latent vectors are forced to distribute on a hyper-spherical surface \zzhu{which} do not follow the actual distribution of shape parameters \zzhu{and make} the reconstruction results inaccurate. \zzhu{On the contrary employing} an intense reconstruction loss \zzhu{would} probably \zzhu{make the distribution of latent vector to be no longer hyperspherical, resulting in less identity-consistent reconstruction results.}  \zzhu{Note that nonlinear face models\cite{tran2018nonlinear,tran2019towards,bagautdinov2018modeling},  which also belongs to the family of 3DMMs,  are not guaranteed to transfer the discrimination of the parameter space to the geometric space as explained in Section\ref{section:my} thus cannot preserve identity information while constructing face models.}

To address the above issue, we propose to keep the latent space of SFMs consistent with that of face recognition. Additionally, it should meet the requirement that the discriminability can be transferred between the shape parameter space and the geometric space. Here, we first introduce the latent space distribution of identity embeddings and then describe how we design the structure of SFMs and the concrete constraints the SFM should satisfy.

\subsection{Hypersphere Manifold of Identity Embedding}
Modern face recognition works always adopt the softmax-based classification loss for metric learning, where weights $w$ and identity embeddings $l$ are normalized and the concept of margin is adopted to further boost discrimination of deep face features. In particular, a loss function with margin can be formulated as
Equation \ref{margin_loss}:
\begin{equation}
\begin{aligned}
    \label{margin_loss}
    &L_{m}=\\&-\frac{1}{m}\sum^m_{i=1}\log \frac{e^{\|x_i\|cos{(\alpha\theta_{y_i}+\beta)-\gamma}}}{e^{\|x_i\|cos{(\alpha\theta_{y_i}+\beta)-\gamma}} + \sum^n_{j=1, j \neq y_i}e^{\|x_i\|cos{(\theta_{j})}}} \\
    & where \qquad cos(\theta_i) = w_il_i
\end{aligned}
\end{equation}
where $x_i \in \mathbb{R}^d$ denotes the $d$ dimensional deep feature of $i$-th sample, $y_i$ denotes the label of  $x_i$.
$\theta_j$ shows the angle between $x_i$ and class $j$ in the feature space. $m$ and $n$ denote the batch size and the class number respectively. The parameter $\alpha$, $\beta$ and $\gamma$ encode the margins of different kinds(ref to SphereFace\cite{liu2017sphereface}, Cosface\cite{wang2018cosface} and Arcface\cite{deng2019arcface}). The identity embedding trained with softmax-based classification are distributed on a hypersphere.

\subsection{Latent Space of Sphere Face Models}
\label{section:my}
As mentioned above, the established SFM should meet the following criteria: (1) the discriminability of the parameter space can be transferred to the discriminability of the geometric space; (2) the distribution of the SFM parameters is consistent with the distribution of the face recognition identity embeddings, that is, the isotropic hyperspherical distribution. 

For the first criteria, SFM parameter space have to meet the following conditions:
\begin{equation}
\label{eq:cond1}
\begin{aligned}
 \forall x_1, x_2, x_3; &   if \left \| x_1 - x_2 \right \| \leq \left \| x_1 - x_3 \right \|  \\  & then \left \| f(x_1) - f(x_2) \right \| \leq \left \| f(x_1) - f(x_3) \right \|
\end{aligned}
\end{equation}
\jdq{If $f(x)$ is a linear function and the basis (mentioned later in Section.~\ref{sec:pipline}) is orthonormal,  the above condition can be met (The property is proved in \cite{jiang2021reconstructing}). Thus we use orthonormal basis in SFM.}

\zzhu{To meet the second criteria, we normalize the shape parameters in SFM. As a consequence, the vector of shape parameters will be constrained on the hypersphere, leading to the cosine angle between two vectors  proportional to their distance in the geometric space. This also brings up a problem that the distance between the result of all human faces and the average face become the same, since all human faces would have the same distance from the origin of the coordinates. Our solution is to add a scalar to control the norm of the face parameters. Similar as \cite{patel2016manifold}, we use scale-normalized shape parameters, namely identity parameters, since they are related to identity information. The scale parameter represents the difference with the mean face.  }

To summarize, our SFM consists of a scale parameter $s$ and a vector of shape parameters $\textbf{x}$ to describe a face model.
\begin{figure*}[ht]
\begin{center}   
\centering 
\includegraphics[width=17cm,trim={4cm 5cm 5cm 5cm},clip]{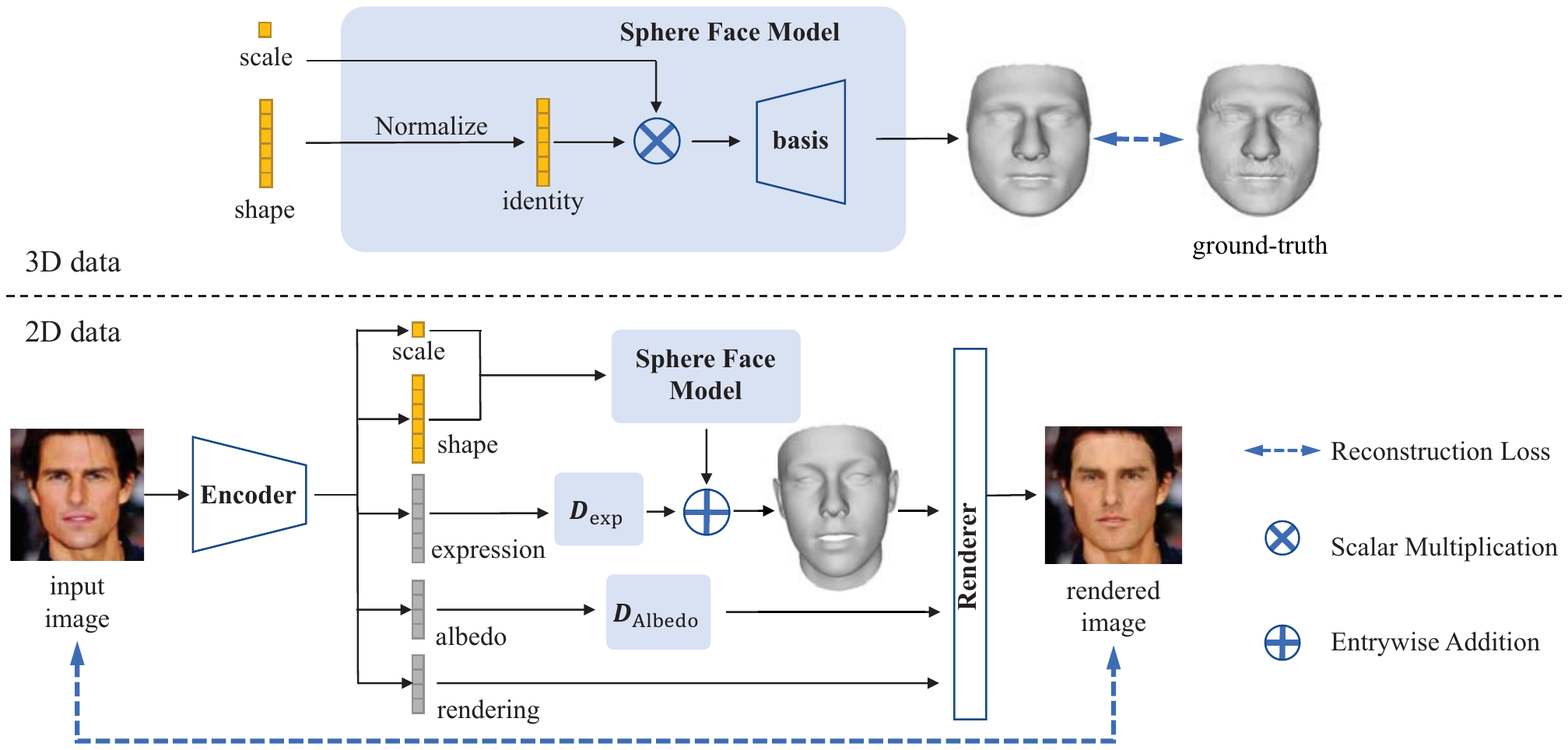} 

\end{center}   
\caption{\textbf{The framework of our method.} \jdq{The normalization of x generates the identity parameters distributing on a hypersphere. The normalized identity parameter is multiplied by the scale parameter to get the shape parameter and goes through the basis to get the corresponding mesh. When training on 3D data, we directly optimize \zy{$s$} and \zy{$x$}. When training on 2D data, we use encoder-decoder because it \zy{requires} other parameters to render the image.}
}
\label{fig:pipeline} 
\end{figure*}

\section{Sphere Face Model}
\label{sec:pipline}
Given the shape parameters $\textbf{x}$ and the scale parameter $s$, our Sphere Face Model is able to reconstruct the 3D face shape by:
\begin{equation}
\mathbf{M} =\mathbf{\bar{M}} + \mathbf{A}(s*\frac{\textbf{x}}{ \left \| \textbf{x} \right \|})
\label{eq:s3dmm} 
\end{equation}
where $\mathbf{M}\in \mathbb{R}^{3n}$ is a reconstructed 3D face shape with $n$ vertices and $\mathbf{\bar{M}}\in \mathbb{R}^{3n}$ is the mean face shape. The normalized term $\textbf{x} / { \left \| \textbf{x} \right \|}$ represents the identity parameters. The orthogonal matrix $\mathbf{A}$ represents the basis of SFM, which is \zy{obtained} by a 2D-3D joint learning framework based on deep neural networks. \jdq{This structure guarantees $s*\frac{\textbf{x}}{\left \| \textbf{x} \right \|}$ located on the hypersphere.}

The previous works for constructing parameterized models mostly rely on either 2D or 3D datasets. However, only training the model with 3D models would lack face variants because there is no publicly available large 3D face datasets. Training only with a two-dimensional dataset is also \zy{difficult} to get satisfying results, since the large diversity of expressions and poses will affect the identity-related features in the reconstructed face models without 3D shape guidance. The regularization constraint used in these methods \cite{tran2018nonlinear,tran2019towards} also makes the generated mesh similar \zy{to} with the average face. Tran et al.\cite{tran2019towards} used the proxy strategy to alleviate that issue, but did not fully solve it. Therefore, we propose an effective learning scheme to utilize both 2D and 3D data to learn face models with aforementioned properties. 

In the following sub-sections, we firstly introduce the overall framework and then describe how the deep model is trained using both 3D and 2D face data.


\subsection{Learning Framework}
Given the model defined in Equation~\eqref{eq:s3dmm}, our goal is to learn the basis matrix $\mathbf{A}$ from face datasets. To achieve so, we adopt a two-stage training strategy as illustrated in Figure \ref{fig:pipeline}. In the first stage we feed the model with scale  and  shape parameters and force the model to reconstruct the 3D face. After this step we obtain a basis matrix, which is  rough  due to the scarcity of the 3D training data. In the second stage we make use of the large 2D face datasets and train an encoder-decoder style model  similar to~\cite{tran2019towards,tewari2018self,zhu2020reda}. The pre-trained SFM can be regarded as a decoder module which can reconstruct the 2D face image along with other decoder modules using the  latent vector from the encoder. By optimizing the encoder and decoder our SFM is fine-tuned. 

More specifically, the encoder regresses the scale parameters, shape parameters, expression parameters, and other parameters for rendering, such as albedo parameters, illumination parameters, pose parameters, and camera parameters. In the decoder part, we have four components, each of which is to be trained in this stage: (1) The trained shape basis of SFM, (2) The expression basis $D_{exp}$ from bfm2017\cite{gerig2018morphable}, (3)the albedo basis $D_{albedo}$  from\cite{smith2020morphable}, (4) the rendering layer takes the geometric, albedo, illumination, pose parameter, and camera parameter and renders 224$\times$224 RGB images, which is based on Pytorch3d\cite{ravi2020pytorch3d}. The illumination model is a spherical harmonic illumination model. 

In previous works, \cite{tran2019towards} did not use 3d prior when constructing face models from 2D data; \cite{tewari2018self} creates a new basis besides the 3DMM to correct face shape; \cite{zhu2020reda} directly regresses the residual displacement in vertex space to correct the face shape. In contrast, our work directly corrects the 3d prior basis by decoupling the expression and appearance information in 2D data, which is able to learn better identity-related features for face reconstruction.



\subsection{Data Preparation}
\textbf{3D data.} FRGC v2.0 database \cite{phillips2005overview} contains 4007 3D face scans of 466 subjects, and is acquired by a Minolta Vivid 900/910 series sensor under controlled illumination conditions. In the preprocessing, we use a non-rigid iterative closest point algorithm\cite{amberg2007optimal} to register the 3D face raw scans to the topology of BFM2017 \cite{gerig2018morphable} and remove the sample with \zzhu{extreme expressions}. The registered 3D models face the positive direction of the z-axis and their centers are coincident with the origin. Note the unit of the registered 3D model is the millimeter.

\textbf{2D data.}
The second stage is trained with 300W-LP\cite{zhu2015high} and VGGFace2\cite{cao2018vggface2}. VGGface2 contains 3.31 million images of 9131 subjects covering a large range of poses, ages, and ethnicities. 300W-LP is a synthetically generated dataset based on the 300-W database\cite{sagonas2016300} containing 61,255 samples across various poses. In our preprocessing stage, the faces are aligned using similarity transformation and cropped to 224$\times$224 in the RGB format with its landmark of 300W-LP.

\subsection{Training Sphere Face Model with 3D Data}
SFMs are first trained \zy{with} 3D data to learn the shape basis using the following loss function:

\textbf{Loss function.}
To assemble the identity parameters of the same identity and separate those of different identities in cosine distance, we apply the softmax loss with normalized shape parameters and normalized weight, which is introduced by \zy{the} Normface\cite{wang2017normface}:
\begin{equation}
L_{m}= -\frac{1}{m}\sum^m_{i=1}\log \frac{e^{\frac{\textbf{x}_{y_i}}{ \left \| \textbf{x}_{y_i} \right \|}*\frac{w_{y_i}}{ \left \| w_{y_i} \right \|}}}{\sum^n_{j=1} e^{\frac{\textbf{x}_j}{ \left \| \textbf{x}_j \right \|}*\frac{w_j}{ \left \| w_j \right \|}}}
\label{loss:norm_softmax} 
\end{equation}
where $n$ is the number of classes and $m$ is the number of samples of the batch. $y_i$ the groundtruth label. $w_j$ represents the $j$th row of the basis $\mathbf{A}$.
\jdq{At the same time, we aggregate the scaled identity parameters $s*\frac{\textbf{x}}{ \left \| \textbf{x} \right \|}$ of the same identity to its center $c$ and separate the centers of different identities in Euclidean distance}:
\begin{equation}
L_{c} = \frac{\left \| s*\frac{\textbf{x}_{y_i}}{ \left \| \textbf{x}_{y_i} \right \|}-c_{y_i} \right \|^2}{\frac{1}{n}\sum_{i \neq j} \left \|c_i-c_j\right \|^2}
\label{loss:cos}
\end{equation}
where $c_i$ represents the center of \zy{the} $i$th class. Finally, we minimize the reconstruction error with basis regularization:
\begin{equation}
L_{s}=  \left \| M - \hat{M} \right \|^2 + \left \| \mathbf{A}^T\mathbf{A} - \mathbf{I} \right \|^2
\label{loss:3d_recon}
\end{equation}
where $M$ is the ground-truth mesh  and $\hat{M}$ is the reconstructed mesh. $I$ is the identity matrix. Finally, the overall loss is:
\begin{equation}
L_{f}=  L_{m} + L_{c} + L_{s} 
\label{loss:3d_all}
\end{equation}

\textbf{Hyperparameter setting.} We use the Adam optimizer, where the initial learning rate of $x$ and $s$ is 0.02 and that of the weight of $\mathbf{A}$ is 0.005. The batch size is 512, and the learning rate is reduced to one-tenth for every 20 epochs.

\subsection{Training Sphere face model with 2D Data}

In the second stage, we train a model that is able to reconstruct the 2D face image. Here, the decoder is initialized by the first stage and will be fine-tuned during this stage. \jdq{Here, $\varepsilon$ denotes the weight of a loss term.}

\textbf{loss function}
\jdq{The loss function consists of three components: landmark loss, photometric loss, and recognition loss. The landmark loss and recognition loss would take effect according to the label of training data as follows:}
\begin{equation}
L =\left\{
\begin{array}{lcl}
L_{pix}(I_{r},I) + \varepsilon_{l} L_{land}(I_{r},I)        &      & {I \in
S_{recon}}\\
 L_{pix}(I_{r},I) + \varepsilon_{s} L_{id}(I_{r},I)        &      & {I \in  S_{id}}
\end{array} \right.
\label{loss_overall}
\end{equation}
\jdq{where $L_{pix}$ is the photometric loss, $L_{land}$ is the landmark loss, and $L_{id}$ is the recognition loss. $I_r$ is the rendered image and $I$ is the input image. The set $S_{recog}$ represents the training data with landmark annotations and $S_{id}$ is the the training data with identity annotations. We explain these losses in detail below.} 

The landmark term $L_{land}$ uses the $L_1$ loss between projected landmarks $\hat{V}_{2d}$ and ground-truth landmarks ${V}_{2d}$:
\begin{equation}
L_{land} = \frac{1}{N} \left \| V_{2d}-\hat{V}_{2d} \right \|_2
\label{loss_proj}    
\end{equation}
where $N$ is the number of landmarks.


Photometric loss measures the difference between the rendered image and the input image using pixel-wise differences \zy{to measure} the absolute errors between each corresponding pixel pair with the weights of a confidence map\cite{wu2020unsupervised}, which aims to deal with occlusions or other challenging appearance variations such as beard and hair. The weighted pixel-wise loss is defined as follows:\jdq{
\begin{equation}
L_{pix}(I_r,I)=-\frac{1}{|\Omega|} \sum_{u v \in \Omega} \ln \frac{1}{\sqrt{2} \sigma_{u v}} \exp -\frac{\sqrt{2} \ell_{1, u v}}{\sigma_{u v}}
\label{pixelLoss}
\end{equation}
where $\ell_{1, u v}=\left|I_r^{u v}-I^{u v}\right|$ is the $L_1$ distance between the intensity of input image $I$ and the reconstructed image $I_r$ at location (u, v) and $\sigma \in \mathbb{R}_{+}^{W \times H}$ is the confidence map. $\Omega$ is the 2D image space.}  

As shown in Equation(\ref{loss_recon}), the regularization term $L_{reg}$ consists of two parts: parameter-level regularity loss $L_{preg}$ and mesh-level regularity loss $L_{mreg}$.  
\begin{equation}
L_{reg} = L_{preg} + \varepsilon_{mreg} L_{mreg}
\label{loss_recon}    
\end{equation}
The regularization term of $L_{regp}$ for 3DMM coefficients is:
\jdq{\begin{equation} 
L_{preg} = \varepsilon_{id}\sum^{m_{id}}_{j=1}\alpha_{id_j}^2+ \varepsilon_{exp}\sum^{m_{exp}}_{j=1}\frac{\alpha_{exp_j}^2}{\sigma_{exp_j}^2} + \varepsilon_{alb}\sum^{m_{alb}}_{j=1}\frac{\alpha_{alb_j}^2}{\sigma_{alb_j}^2}
\label{para_regular}
\end{equation}
where $\sigma_{exp}$ is an eigenvalue of the expression basis and $\sigma_{alb}$ is an eigenvalue of the albedo basis. $\alpha_{id}$, $\alpha_{exp}$ and $\alpha_{alb}$ are the 3DMM parameters which are regressed by the encoder network as show in Figure \ref{fig:pipeline};  $m_{id}$, $m_{exp}$ and $m_{alb}$ are the dimensions of the shape, expression and albedo parameters respectively.}

The mesh-level regular loss consists of the smooth loss, the symmetrical loss and the residual loss.
\begin{equation}
\label{eq:mreg}
L_{mreg} = L_{smooth} + L_{sym} + L_{res}
\end{equation}

\begin{equation}
\begin{aligned}
L_{smooth}(\mathbf{G})= \frac{1}{N}\sum_{i=1}^N \left\|\mathbf{G}_i-\frac{1}{\left|\mathcal{N}_{i}\right|} \sum_{\mathbf{G}_{j} \in \mathcal{N}_{i}} \mathbf{G}_j\right\|_{2} 
\end{aligned}
\label{albedo_regular}
\end{equation}
where $G$ is the reconstructed face shape, $\mathcal{N}_{i}$ denotes a set of a neighboring vertices $\mathbf{G}_{i}$ and $N$ is the number of vertices.

\jdq{We assume that the human faces in natural expressions are symmetric about the center axis and add the face shape geometry symmetrical loss}:
\begin{equation}
L_{sym}(\mathbf{G}) =  \left \| \mathbf{G} -filp(\mathbf{G}) \right \|_1 
\end{equation}
where $flip()$ is the operation to flip the face shape geometry.

The residual loss is:
\begin{equation}
    L_{res}(\mathbf{G}) =   \left \| \mathbf{G} - \mathbf{\bar{G}} \right \|_1 
\end{equation}
where $\mathbf{\bar{G}}$ is the mean face geometry.

Face recognition loss includes three components as shown in Equation\eqref{loss_recog}: a softmax-based loss, a centerness loss and a Kullback-Leibler loss.

\begin{equation}
L_{id} =  L_{soft} + \varepsilon_{center} L_{center}  + \varepsilon_{kl} L_{kl}
\label{loss_recog}
\end{equation}
We use cosloss\cite{wang2018cosface} $L_{soft}$ as the softmax-based loss, which applies to the identity parameters. The Kullback-Leibler loss \cite{kingma2013auto} $L_{kl}$ and $L_{center}$ center loss\cite{wen2016discriminative} are applied to the scale parameter.

\textbf{More Training Details}
\zy{Currently}, there are no large public databases that contain both face identity labels and landmark labels. Moreover, since the results of existing face detectors are unsatisfactory in challenging conditions, we do not automatically generate landmarks in the face recognition dataset. Therefore, we use the mixed data from 300W-LP\cite{zhu2015high} and VGGFace2\cite{cao2018vggface2}. To successfully train our model with the mixed dataset, we use the following strategy to achieve convergence: 

(1)\textit{Switch the loss function: } Because the labels in the mixed database are deficient, we determine which loss terms take effect according to the labels of the training samples. For example, if the training sample is from VGGface2, we enable the face recognition loss and photometric loss. Otherwise, the landmark loss and photometric loss take effect as shown in Equation(\ref{loss_overall})

(2) \textit{Warm up the network: }To warm up the network, we train our network on the 300W-LP\cite{zhu2015high} database only using $S_{recon}$, then train the mixed database with the whole loss function shown in \ref{loss_overall}.

(3) \textit{Balance the data from different datasets:}
Because the VGGface2 contains 3.31 million images while 300W-LP\cite{sagonas2016300} contains 61,255 samples, which are extremely unbalanced, we design a sampling scheme where the probability of selecting samples from the VGGFace2 is given by:
\begin{equation}
P = \frac{N_{recon}}{N_{recog}+N_{recon}}
\end{equation}
Here, $N_{recon}$ is the number of samples in 300W-LP dataset and $N_{recog}$ is that in VGGFace2 dataset. The probability of selecting samples from 300W-LP database is $1-P$.

\begin{table}
\centering
\begin{tabular}{l|cccccc}
\toprule
Model & PCA & Linear & Sphere-Linear  & SFM \\
\midrule 
RMSE & \textbf{0.2777} & 0.2916 & 0.2808  & 0.2827 \\
SCE & -0.0490 & -0.0492 & -0.0490  &  \textbf{0.1193} \\
SCC & -0.1068 & -0.1073 & -0.1068  & \textbf{0.2038}  \\
CH & 15.86 &  15.89 & 15.86  &  \textbf{25.81} \\
\bottomrule 
\end{tabular}
\caption{The results of model representation ability and its shape parameter separability in FRGCv2 database.}
\label{tab:frcg_3d}
\end{table}

\begin{table}
\centering
\begin{tabular}{l|cccccc}
\toprule
Model & PCA & Linear & Sphere-Linear  & SFM \\
\midrule 
RMSE & \textbf{0.3747} & 0.3924 & 0.3790  & 0.3863 \\
SCE & 0.1061 & 0.1059 & 0.1061  &  \textbf{0.2236} \\
SCC & 0.1474 & 0.1470 & 0.1473  & \textbf{0.3513}  \\
CH & 9.01 &  9.03 & 9.01  &  \textbf{10.28} \\
\bottomrule 
\end{tabular}
\caption{The results of model representation ability and its shape parameter separability in Bosphorus database.}
\label{tab:bos_3d}
\end{table}

\begin{table}
\centering
\begin{tabular}{l|cccccc}
\toprule
Model & BFM17 & FLAME   & SFM &  SFM* &\\
\midrule 
RMSE & 0.6137  & 0.4820   & \textbf{0.3501} & 0.4355\\
SCE & 0.1674 & 0.0928   &  0.2247 &  \textbf{0.2953} \\
SCC &  0.2451 & 0.1700   & 0.3616 &  \textbf{0.4514} \\
CH & 6.87  &   3.85  &  9.50  &  \textbf{11.04} \\
\bottomrule 
\end{tabular}
\caption{Compare the model representation ability and its shape parameter separability with BFM17\cite{gerig2018morphable} and FLAME\cite{li2017learning} in Bosphorus database.The SFM* is the SFM finetuned with 2d data.}
\label{tab:3d}
\end{table}

\begin{figure}[t]
\begin{center}
\centering 
\includegraphics[width=9cm,trim={1cm 5cm 5cm 6cm},clip,angle=0]{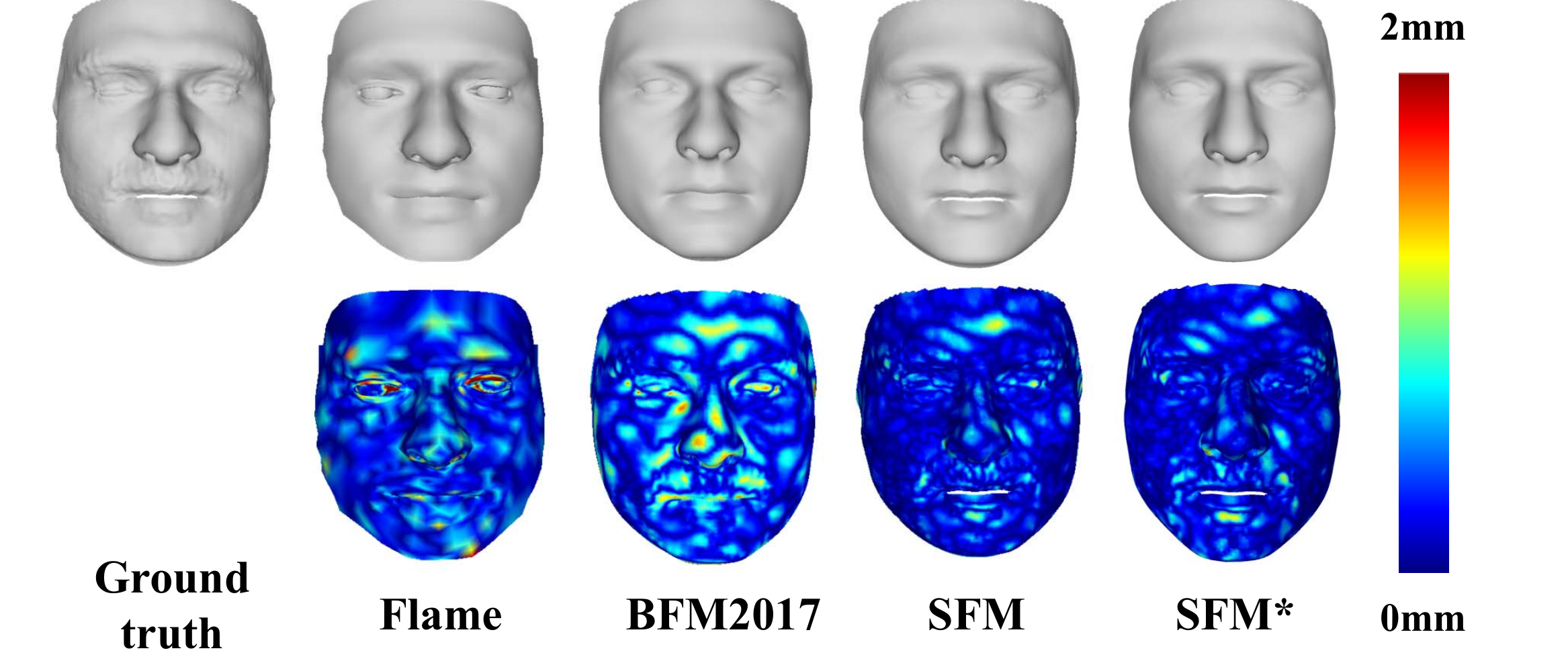} 
\end{center}   
\caption{The fitting results of BFM17\cite{gerig2018morphable}, FLAME\cite{li2017learning} and ours. The first row is the fitted mesh and the second row is the error map with ground truth. SFM* is the SFM fine-tuned with 2d data.}
\label{fig:3derror_map}
\end{figure}

\begin{figure}[t]
\begin{center}
\centering 
\includegraphics[width=9cm,trim={0.5cm 4cm 2cm 4cm},clip,angle=0]{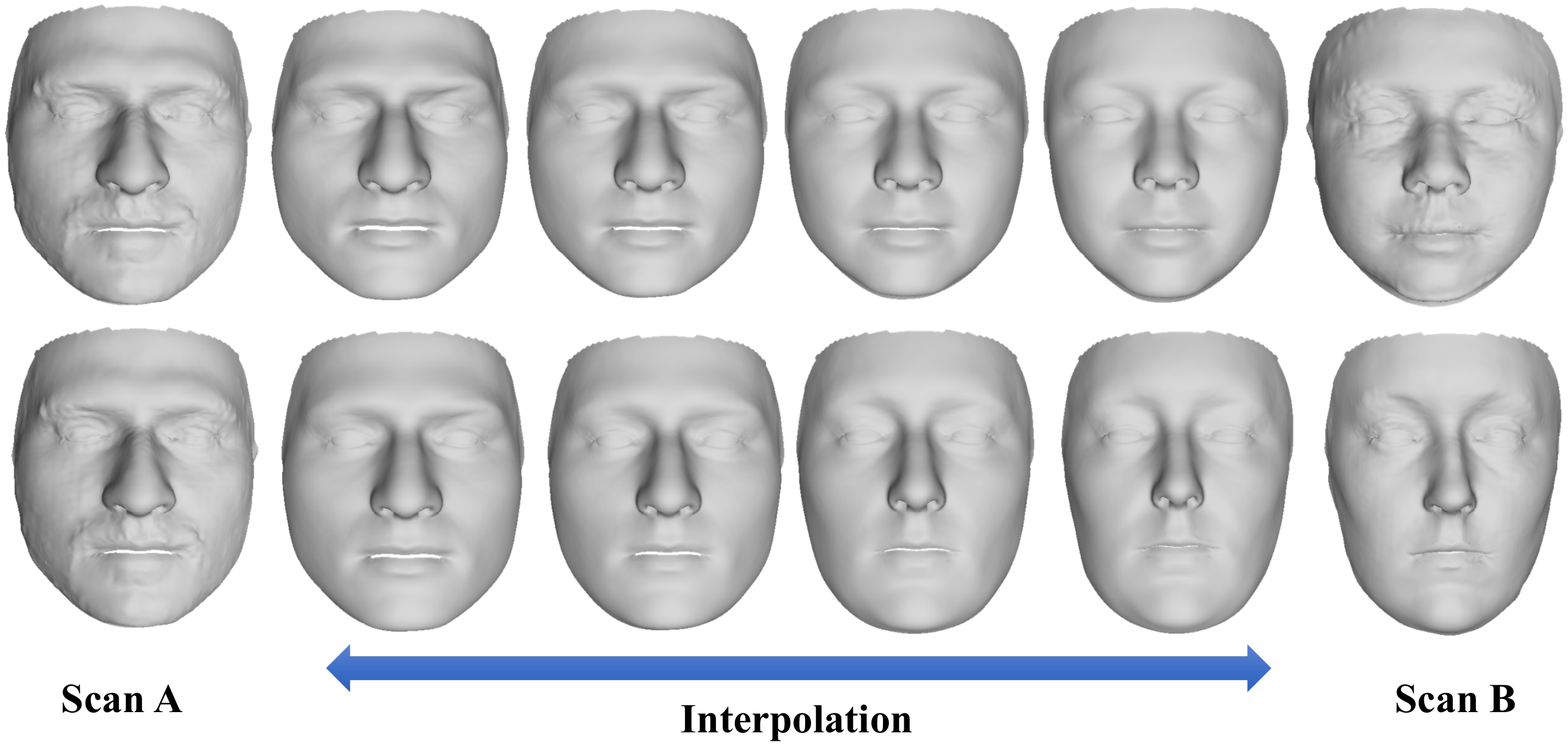} 
\end{center}   
\caption{The latent vector distributions of different methods. We select 20 people on FRGCv2, fitting the shape parameter, and then use t-SNE to reduce the shape parameter to two dimensions and display it on this figure. Different colors represent different people.}
\label{fig:interpolation}
\end{figure}

\begin{figure}[t]
\begin{center}
\centering 
\includegraphics[width=9cm,trim={4cm 4cm 5cm 5cm},clip,angle=0]{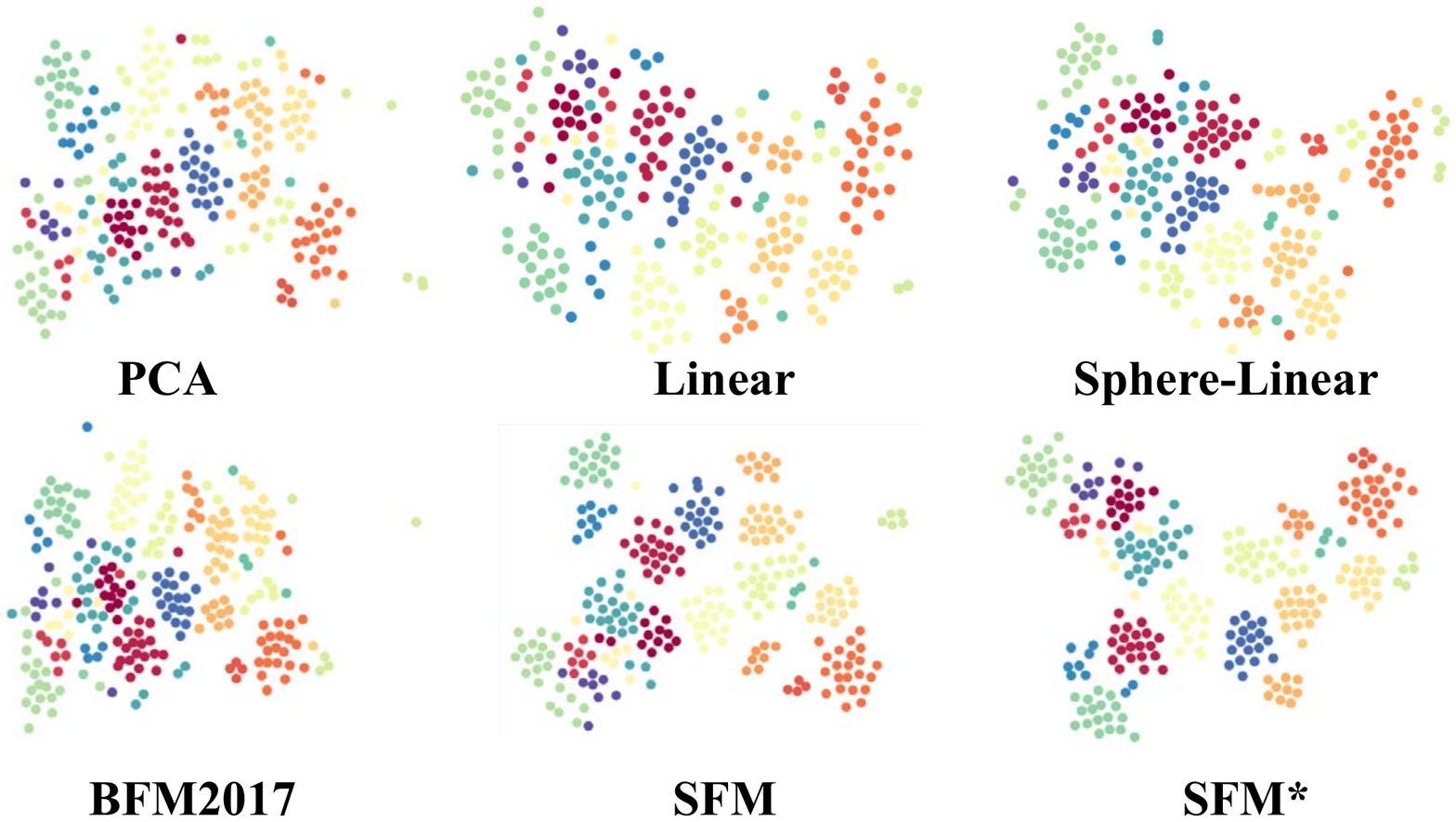} 
\end{center}   
\caption{The latent vector distributions of different methods. We select 20 people on FRGCv2, fitting the shape parameter, and then use t-SNE to reduce the shape parameter to two dimensions and display it on this figure. Different colors represent different people.The SFM* is the SFM fine-tuned with 2d data.}
\label{fig:d_latent}
\end{figure}

\begin{table}
\begin{center}
\begin{tabular}{|l|c|c|c|}
\hline
Method & LFW & CFP-FP & YTF    \\
\hline
\multicolumn{4}{|c|}{Cosine similarity} \\
\hline
3DMM-CNN  & 90.53  & - & 88.28 \\
Lui et al. & 94.40 & - & 88.74 \\ 
D3FR &88.98 & 66.58 & 81.00 \\
TDDFA & 64.90 & 57.57 & 58.50 \\
MGCNet &82.10 & 70.87 & 75.58 \\
RingNet &79.40 & 71.41 & 71.02 \\
Jiang et al & 95.36 & 83.34 & 89.07 \\
SFM & \textbf{98.23} & \textbf{91.12} & \textbf{93.86} \\
\hline
\multicolumn{4}{|c|}{Euclidean similarity} \\
\hline
D3FR &87.63 & 66.50 & 81.10 \\
TDDFA & 63.45 & 55.49 & 58.16 \\
MGCNet &80.87 & 66.01 & 72.36 \\
RingNet &80.05 & 69.46& 72.40 \\
Jiang et al & 94.47 & 80.78  & 86.40 \\
SFM & \textbf{98.03} & \textbf{90.79}  &  \textbf{92.60} \\
\hline

\end{tabular}
\end{center}
\caption{Face verification accuracy(\%) on the LFW, CFP-FP and YTF datasets. Our results are obtained using the  weighted center loss. We compare our results with 3DMM-CNN\cite{tuan2017regressing}, Liu et al.\cite{liu2018disentangling}, D3FR\cite{deng2019accurate}, TDDFA\cite{guo2020towards}, MGCNet\cite{shang2020self}, Jiang et al\cite{jiang2021reconstructing}  and RingNet\cite{sanyal2019learning}.}
\label{tab:FG_acc2}
\end{table} 

\section{Experiment}
Comparing with the previous methods, SFMs have the following properties: (1) The latent space of SFMs has inherent separation property between the various classes; (2) The latent space distribution of SFM is similar to that of identity embeddings, so that the losses for face recognition and face reconstruction can be easily optimized together in the pipeline of shape-consistent face reconstruction; (3) SFM has better capabilities for face representation. Therefore, in this section, we evaluate SFMs from the following three aspects: model representation ability, separability of latent space, and the performance of shape-consistent monocular reconstruction.

\begin{figure}[t]
\begin{center}
\centering 
\includegraphics[width=9cm,trim={1cm 4cm 18cm 4cm},clip,angle=0]{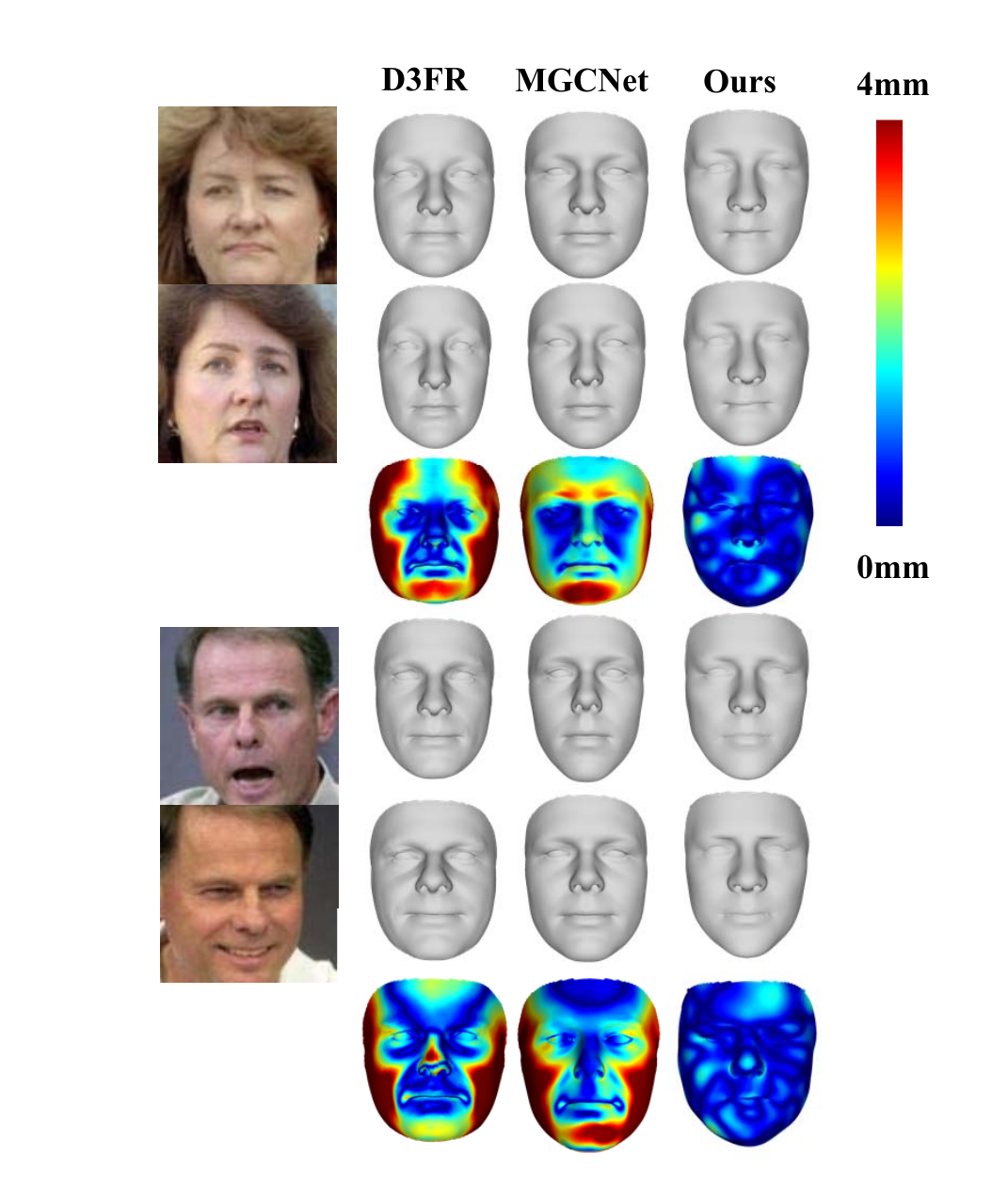} 
\end{center}   
\caption{Comparison with D3FR\cite{deng2019accurate}, MGCNet\cite{shang2020self} on LFW samples
The reconstruction results are the same person under different conditions. The third and sixth rows are the error between two meshes in the same column.}
\label{fig:vis_s}
\end{figure}

\begin{figure}[t]
\begin{center}
\centering 
\includegraphics[width=9cm,trim={2cm 4cm 10cm 4cm},clip,angle=0]{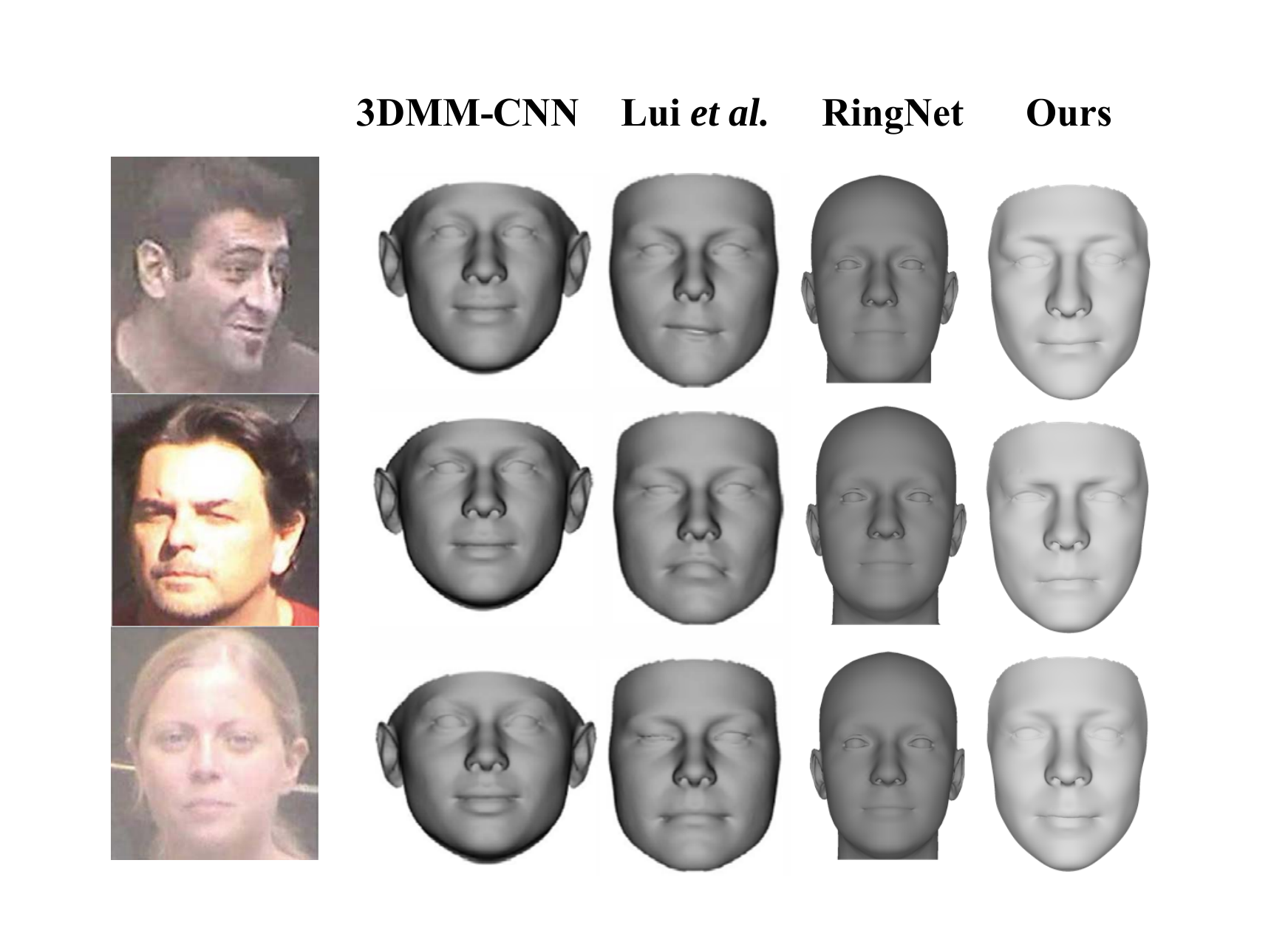} 
\end{center}   
\caption{Comparison with 3DMM-CNN\cite{tuan2017regressing}, liu et al.\cite{liu2018disentangling} and Ringnet\cite{sanyal2019learning} on three MICC subjects. Our reconstruction results capture more face details.}
\label{fig:vis_f}
\end{figure}

\begin{figure}[t]
\begin{center}
\centering 
\includegraphics[width=9cm,trim={0cm 8cm 18cm 4cm},clip,angle=0]{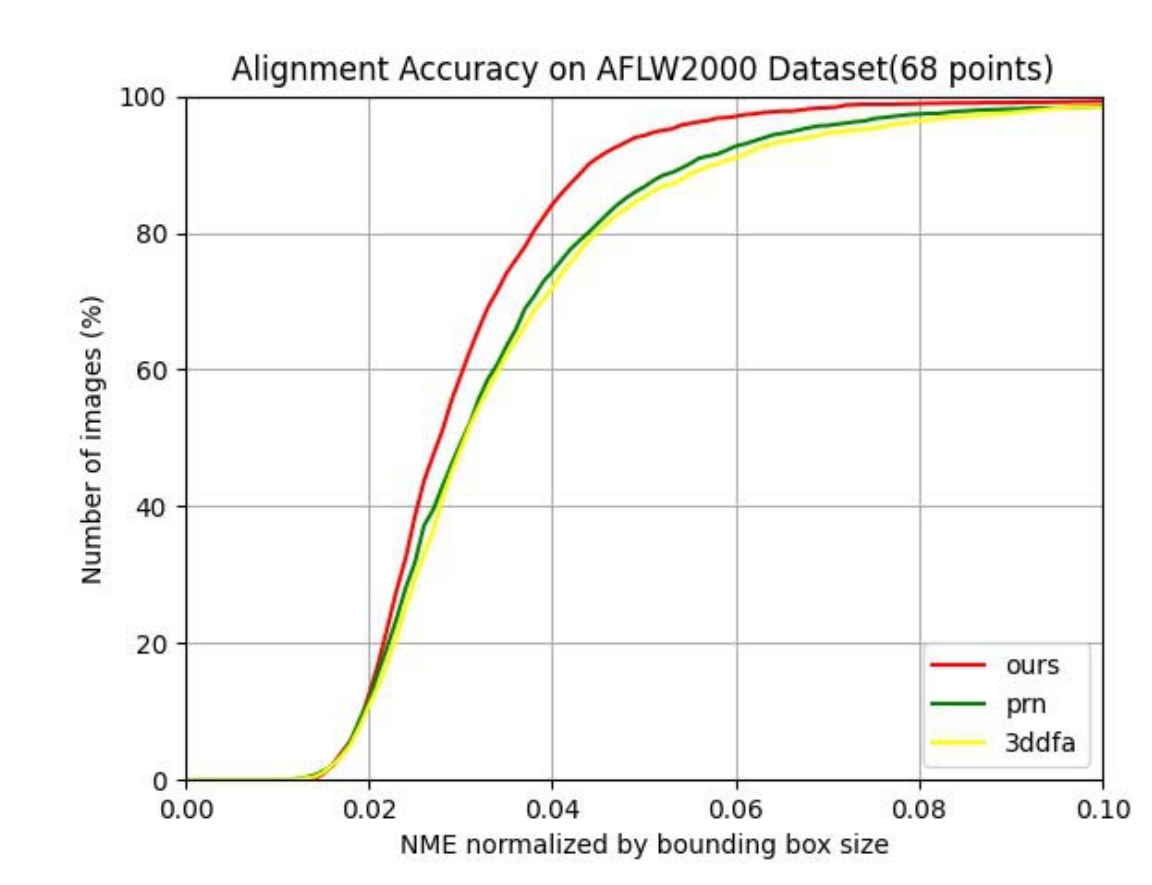} 
\end{center}   
\caption{The cumulative errors distribution curve of 3d face alignment accuracy on \zy{the} ALFW2000 Dataset. Compared with PRNet\cite{feng2018joint}, 3ddfa\cite{zhu2017face}, our method produces \zy{better} results.}
\label{fig:ced}
\end{figure}
 
\begin{figure*}[t]
\begin{center}
\centering 
\includegraphics[width=17cm,trim={0cm 4.5cm 0cm 6cm},clip,angle=0]{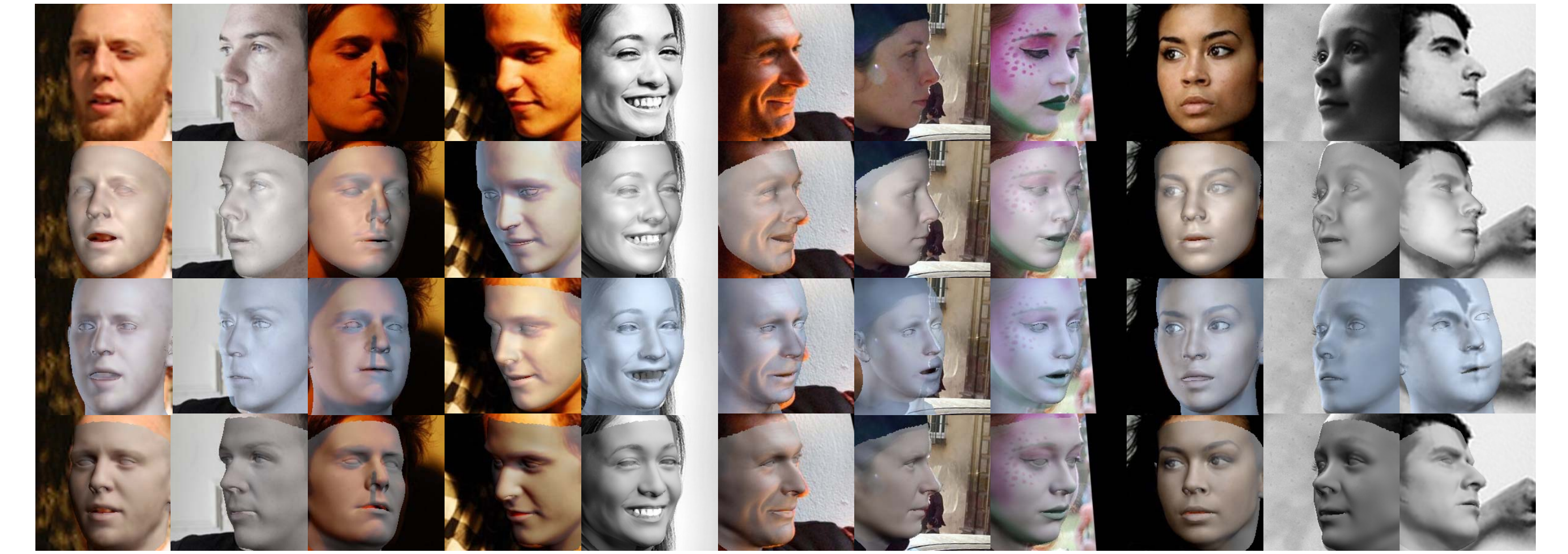} 
\end{center}   
\caption{The visualization results on ALFW2000 dataset. The first row: images from  ALFW2000 dataset. The second row: the result of 3DDFA v2\cite{guo2020towards}. The third row: the results of ringnet\cite{sanyal2019learning}. The forth row: the results of ours.}
\label{fig:alignment}
\end{figure*}

\begin{figure}[t]
\begin{center}
\centering 
\includegraphics[width=9cm,trim={0cm 0cm 0cm 0cm},clip,angle=0]{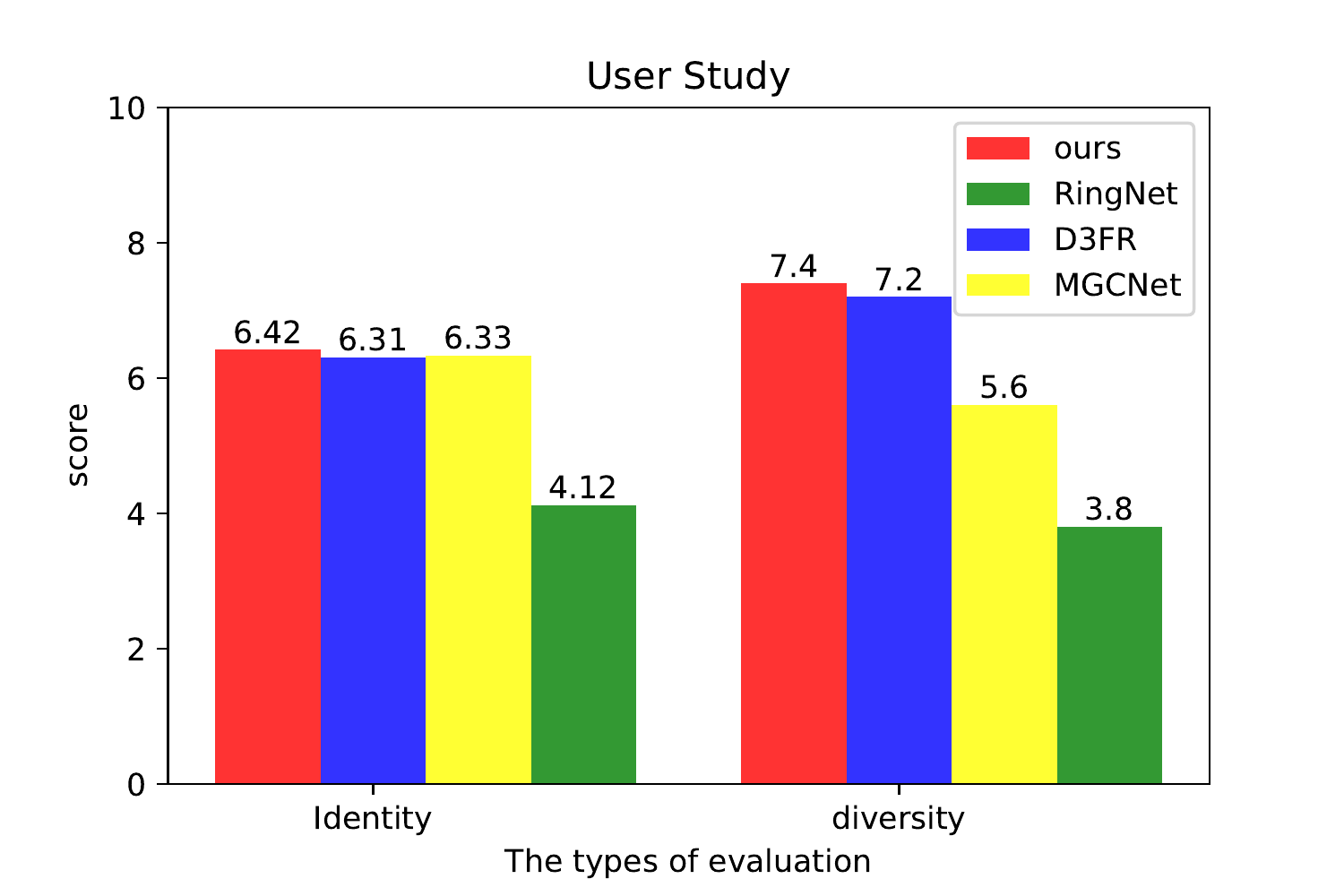} 
\end{center}   
\caption{The results of user study. We compare our SFM with RingNet\cite{sanyal2019learning}, D3FR\cite{deng2019accurate}, MGCNet\cite{shang2020self} \zy{in terms of} identity and diversity, \zy{and our results are more satisfactory}.}
\label{fig:user_s}
\end{figure}

\begin{figure*}[t]
\begin{center}
\centering 
\includegraphics[width=17cm,trim={0cm 2cm 0cm 0cm},clip,angle=0]{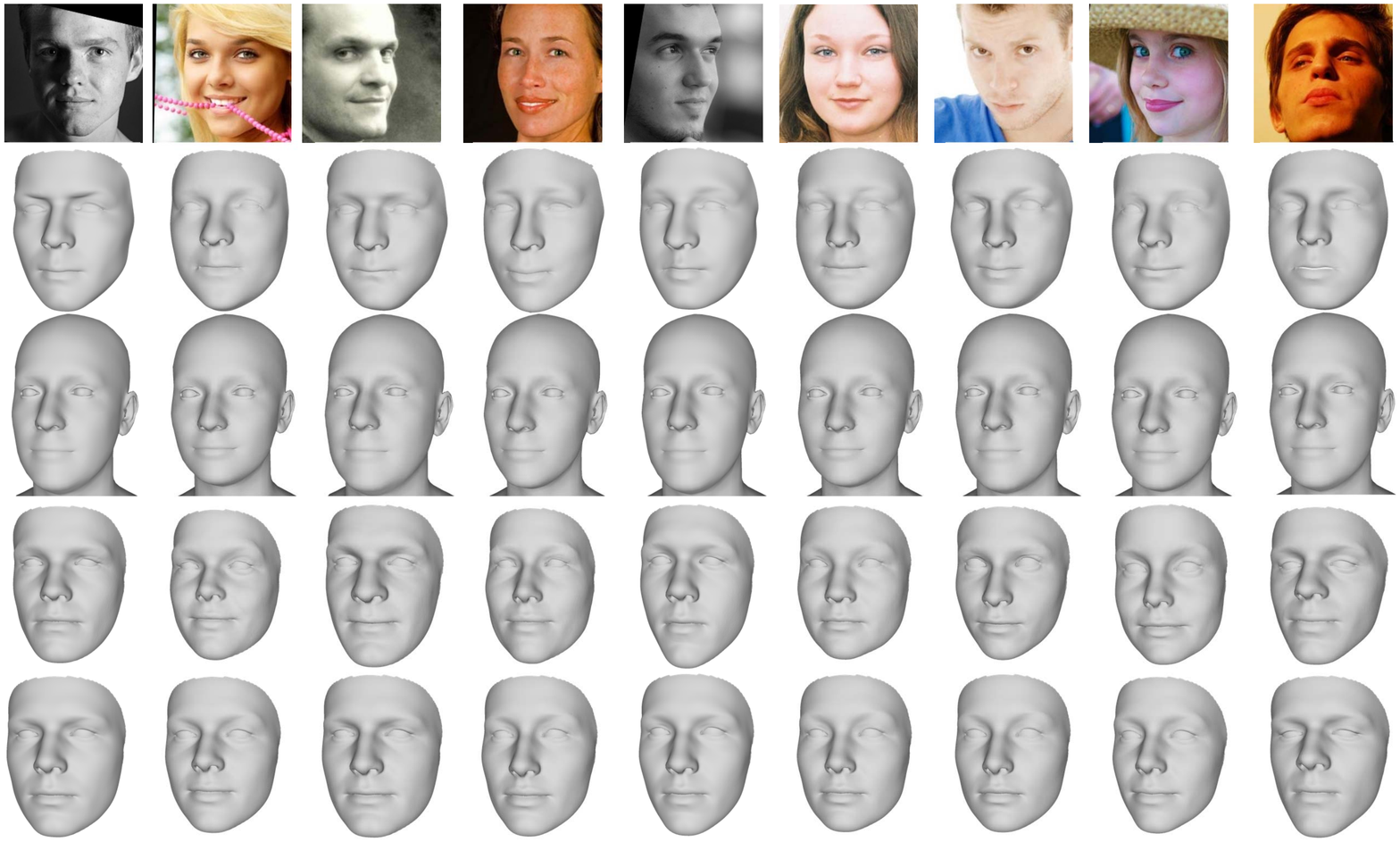} 
\end{center}
\caption{Some samples of user study, The first row: samples from ALFW2000 dataset. The second row: our results. The third row: the results of ringnet\cite{sanyal2019learning}. The forth row: the results of D3FR\cite{deng2019accurate}. The last row: the results of MGCNet\cite{shang2020self}.}
\label{fig:user}
\end{figure*}

\subsection{Model Representation Ability}

To validate the expressive ability of face representation models, we reconstruct 3D meshes on the training and testing database respectively with the same dimension of the latent vector(all of 199 dimensions in this paper). Evaluation on the training database shows the ability of the models to recover the meshes of the training data. We also verify the generalizability of our model by fitting meshes for the testing database. We also present the result of the parameter interpolation.

Our training dataset is FRGCv2 \cite{phillips2005overview}, and the testing dataset is the Bosphorus Database \cite{savran2008bosphorus}, which contains 4666 3D face models of 105 people. The models for each person have various expressions, poses, and occlusions. In our experiment, we select the face with a frontal natural expression for each person, and register all the data on the BFM 2017\cite{bagautdinov2018modeling} template using non-rigid ICPs \cite{amberg2007optimal}. We use the Adam optimizer to optimize the face parameters of the model. The initial learning rate is 0.02 and reduced by a factor of 0.5 for every 128th iteration. The total optimization iteration number is 1000. 

The Root Mean Square Error(RMSE)  between the reconstructed meshes and the ground truth for the training dataset is shown in Table \ref{tab:frcg_3d}, and that for the testing dataset is in Table \ref{tab:bos_3d}.
When generating the above results, we use the face model trained with FRGCv2 but using different methods. ``PCA'' means the face model is directly established by the PCA method. ``Linear'' means the face model is established by optimizing an orthogonal linear basis. ``Sphere-Linear'' refers to using the structure of SFM without the loss of face recognition when constructing the face shape. The expression ability of our SFM basis is slightly better than that of the linear basis but worse than PCA. Because when the face model has a linear orthogonal basis, the basis solved by PCA has the smallest reconstruction error, which is the optimal solution. Our reconstruction accuracy is slightly lower than that of PCA, but has a better separation in the parameter space.

Table \ref{tab:3d} shows the comparison between SFMs with the shape models of BFM2017 and FLAME on the Bosphorus database. We crop the face area for fitting because other areas (ears, neck) are irrelevant to our task and can largely influence the RMSE. We use the point-to-plane error to calculate RMSE. The results show that our face model has less reconstruction errors than others. Figure \ref{fig:3derror_map} shows some fitting results on the Bosphorus database. SFMs are competitive among all the validated 3DMMs in terms of expressive ability, with the best visual quality of the generated reconstruction results.

Figure \ref{fig:interpolation} shows that the parameters of our basis have a good linear interpolation performance. We use the geodesic distance to interpolate the identify parameters and directly interpolate the scale parameters linearly.

\subsection{Separability of Latent Space}
After fitting all the 3D scans of a database, we get the parameters of the corresponding 3DMM model in this database. We can evaluate the clustering properties of these parameters to estimate the degree of separation of these latent space parameters. The performance of clustering can be evaluated with the following metrics: the Silhouette Coefficient score with Euclidean distance(SCE), Silhouette Coefficient with Cosine distance(SCC), and Calinski-Harabasz score indicators(CH). The Silhouette Coefficients is given as:
\begin{equation}
s = \sum^n_{i=1} \frac{a_i-b_i}{max(a_i,b_i)}
\label{metric:Silhouette Coefficient}
\end{equation}
where $a_i$ is the mean distance between $i$th sample and all other points of the same class and $b_i$ is the mean distance between $i$th sample and all other points of the nearest cluster. $n$ is the number of the sample. The score is the ratio of the sum of between-cluster dispersion and of within-cluster dispersion for all clusters (where dispersion is defined as the sum of squared distances ).  The Calinski-Harabasz score(CH) is defined as the ratio of the between-clusters dispersion mean and the within-cluster dispersion: 
 \begin{equation}
     s = \frac{tr(\mathbf{B}_k)}{tr(\mathbf{W}_k)} \times \frac{n_E-k}{k-1}
 \end{equation}
where 
$\mathbf{B}_k$ is the trace of the between-cluster dispersion matrix and $\mathbf{W}_k$  is the trace of the within-cluster dispersion matrix defined by:
\begin{equation}
\begin{aligned}
     & \mathbf{W}_k = \sum^k_{q=1}\sum_{x \in C_q}(x-c_q)(x-c_q)^T \\
    & \mathbf{B}_k = \sum^k_{q=1}(n_q)(c_q-c_e)(c_q-c_E)^T
\end{aligned}
\end{equation}
with $C_q$ the set of points in cluster $q$, $c_q$ the center of cluster $q$, $c_E$ the center of $E$, and $n_q$ the number of points in cluster $q$.

Table\ref{tab:frcg_3d}  and Table \ref{tab:bos_3d} show the results of shape parameter space separation of SFMs and the shape basis constructed by other methods. We add the face recognition loss while establishing the SFM basis, which significantly improves the parameter space separation. Table \ref{tab:3d} shows the comparison between our basis and other basis. The separability of our parameter space is also higher than other models.
In order to present the distribution of latent space more intuitively, we use t-SNE\cite{van2008visualizing} to project the shape parameters of different bases to two dimensions. As shown in Figure \ref{fig:d_latent}, the intra-class distance of the latent space of SFM is small, and the inter-class distance is large. Compared with other methods, the latent space of our basis shows a much better separation.

\subsection{Monocular Reconstruction}
\zy{To} test the face monocular reconstruction, we use the same encoder-decoder network in the second training stage as the shape-consistent face reconstruction pipeline. Unlike the training phase, when performing inference for monocular reconstruction, we fix the weight of the shape basis and retrain the encoder to regress the parameters. In this section, we evaluate the faithfulness and shape consistency of monocular face reconstruction results using SFM. In terms of faithfulness, we compared the visual results with other face reconstruction methods. Moreover, we compare the accuracy of 3D face alignment. In terms of shape consistency, we compare the accuracy of the face recognition using the shape parameters and the visual results of the same person reconstructed in different environments. In this subsection, when comparing with methods, ``ours'' means that we use the SFM finetuned with 2D data.

\textbf{Face shape consistency evaluation.}
We use the cosine distance and Euclidean distance as the measurements of the similarity between two groups of shape parameters, when evaluating the face recognition accuracy. The result of face recognition performance is shown in Table \ref{tab:FG_acc2}. The accuracy of our face recognition parameters is higher than other methods. Figure \ref{fig:vis_s} shows the visualization results of the 3D face reconstructed by the same person in different environments. We have smaller errors among the meshes reconstructed for the same person.

\textbf{Face faithfulness evaluation.}
Figure \ref{fig:vis_f} shows \zy{that} our reconstruction results capture more face details compared to other face reconstruction methods. Figure \ref{fig:ced} shows the cumulative errors distribution curve of 3D face alignment compared with other methods, \jdq{Figure \ref{fig:alignment} shows the visual results of face alignment and Figure \ref{fig:user} shows the visual results of face shape.}
\zy{Both quantitative and visual evaluations show that \zzhu{in terms of face faithfulness} our method has better performance than previous methods.}

\textbf{User study.}
We conducted a user study to compare the visual diversity and the degree of retention of the reconstructed face shape on the identity information. We randomly selected 20 face images from ALFW2000 and reconstructed 3D face models using the following methods: RingNet\cite{sanyal2019learning}, D3FR\cite{deng2019accurate}, MGCNet\cite{shang2020self} and our SFM, and in turn asked 5 participants to evaluate the reconstructed faces' diversity and the retention of identity information of the reconstructed faces from the input image with a score from 0 to 10. Participants were told that the reconstruction results with more identity information maintained or more diversity of different people should be scored higher. The average scores of the results from different methods are shown in the Figure \ref{fig:user_s}. The ``identity'' means the degree of identity preservation and the ``diversity'' means the diversity among the 3D faces reconstructed from different people. Results and comparison vividly show the advantages of our method.

\section{Conclusion}
\zzhu{We have proposed a novel 3D morphable model with a hypersphere manifold latent space for face generation. We have also proposed a two-stage training framework where both 3D and 2D data were utilized. Our model outperformed previous models on the consistency and the fidelity of the reconstructed faces. Experimental results validated that our method is superior to previous methods objectively, and user study showed that our model can provide visually better face reconstruction results. }


{\small
\bibliographystyle{cvm}
\bibliography{cvmbib}
}

\end{document}